\documentclass[letterpaper, 10 pt, conference]{ieeeconf}  

\IEEEoverridecommandlockouts                              

\usepackage{tikz}
\usepackage{xparse}

\tikzset{
  annotatedImage/x/.initial = 0.7,
  annotatedImage/y/.initial = 0.7,
  annotatedImage/width/.initial = 1,
  annotatedImage/.unknown/.code = {
    \edef\tikzappend{\noexpand\tikzset{annotatedImage/.append style =
                {\pgfkeyscurrentname=\pgfkeyscurrentvalue}}}
    \tikzappend
  },
  annotatedImage/.style = {
  }
}

\newsavebox\annotatedImageBox

\newcommand\AnnotatedImageVal[1]{\pgfkeysvalueof{/tikz/annotatedImage/#1}}
\newcommand\SetUpAnnotatedImage[2]{
    \tikzset{annotatedImage/.cd, #1}%
    \sbox\annotatedImageBox{\includegraphics[width=\AnnotatedImageVal{width}\textwidth,
                                          keepaspectratio]{#2}}%
    \pgfmathsetmacro\annotatedHeight{\ht\annotatedImageBox/28.453}
    \pgfmathsetmacro\annotatedWidth{\wd\annotatedImageBox/28.453}%
}

\NewDocumentCommand\annotatedImage{ O{} m m}{%
  \bgroup
    \SetUpAnnotatedImage{#1}{#2}%
    \begin{tikzpicture}[xscale=\annotatedWidth, yscale=\annotatedHeight]%
        \node[inner sep=0, anchor=south west] (image) at (0,0) {\usebox{\annotatedImageBox}};
        \node[annotatedImage] at (\AnnotatedImageVal{x},\AnnotatedImageVal{y}) {#3};
    \end{tikzpicture}
  \egroup%
}

\newcommand\annotate[1][]{\node[annotatedImage,#1]}

\newenvironment{AnnotatedImage}[2][1]{%
  \SetUpAnnotatedImage{#1}{#2}%
  \tikzpicture[xscale=\annotatedWidth, yscale=\annotatedHeight]
    \node[inner sep=0, anchor=south west,inner sep=0] at (0,0) {\usebox{\annotatedImageBox}};
}{\endtikzpicture}

\usepackage{booktabs}
\usepackage{subcaption}
\usepackage{graphicx}

\usepackage[percent]{overpic}
\usepackage{soul}
\usepackage{float}
\usepackage{cite}
\usepackage{multirow} 
\usepackage{amsmath}
\usepackage{amssymb}
\usepackage[acronym]{glossaries}
\usepackage{cleveref}
\usepackage{breqn}
\usepackage{dsfont}
\usepackage{url}

\definecolor{dgreen}{rgb}{0,0,0}
\definecolor{dyellow}{rgb}{.7,.7,0}
\definecolor{dred}{rgb}{1,0,0}
\definecolor{dblue}{rgb}{0,0,0.7}
\definecolor{dorange}{rgb}{0.9,0.5,0.1}

\newacronym{tcn}{TCN}{Time-Contrastive  Networks}
\newacronym{asn}{ASN}{Adversarial Skill Networks}
\newacronym{ppo}{PPO}{Proximal Policy Optimization}
\newacronym{gan}{GAN}{Generative Adversarial Networks}
\newacronym{vae}{VAE}{Variational Auto-Encoder}
\newacronym{rnntcn}{mfTCN}{mfTCN}
\newacronym{fc}{FC}{Fully Connected}

\title{Adversarial Skill Networks: Unsupervised Robot Skill Learning from Video}

\author{
 Oier Mees$^\ast$, Markus Merklinger$^\ast$, Gabriel Kalweit, Wolfram Burgard
\thanks{$^\ast$These authors contributed equally. All authors are with the University of Freiburg, Germany. Wolfram Burgard is also with the Toyota Research Institute, Los Altos, USA. This work has  been supported partly by the Freiburg Graduate School of Robotics and the German Federal Ministry of Education and Research (BMBF) under contract number 01IS18040B-OML.}
}

\begin{document}
\maketitle
\thispagestyle{empty}
\pagestyle{empty}

\begin{abstract}
Key challenges for the deployment of reinforcement
learning (RL) agents in the real world are the discovery, representation and reuse of skills in the absence of a reward function. To this end, we propose a novel approach to learn a task-agnostic skill embedding space from unlabeled multi-view videos.  Our method
learns a general skill embedding independently from the task context by using an adversarial loss. We combine a metric learning loss, which utilizes temporal video coherence to learn a state representation, with an entropy-regularized adversarial skill-transfer loss. The metric learning loss learns a disentangled  representation by attracting simultaneous viewpoints of the same observations and repelling visually similar frames from temporal neighbors. The adversarial skill-transfer loss enhances re-usability of learned skill embeddings over multiple task domains. We show that the learned embedding enables training of continuous control policies to solve novel tasks that require the interpolation of previously seen skills.
Our extensive evaluation with both simulation and real world data demonstrates the effectiveness of our method in learning transferable skills from unlabeled interaction videos and composing them for new tasks. Code, pretrained models and dataset are available at  \url{http://robotskills.cs.uni-freiburg.de}
\end{abstract}



\section{Introduction}
	
Intelligent beings have the ability to discover, learn and transfer skills without supervision. Moreover, they can combine previously learned skills to solve new tasks. This stands in contrast to most current ``deep reinforcement learning'' (RL) methods, which, despite recent progress~\cite{hessel2018rainbow, barth2018distributed, silver2018general}, typically learn solutions from scratch for every task and often rely on manual, per-task engineering of
reward functions. Furthermore, the obtained policies
and representations tend to be task-specific and generally do not transfer  to new tasks.

The design of reward functions that elicit the desired agent behavior is especially challenging for real-world tasks, particularly when the state of the environment might not be accessible. Additionally, designing a reward  often requires the installation of specific sensors to measure as to whether the task has been executed successfully~\cite{schenck2017visual,rusu2017sim}. In many scenarios, the need for  task-specific engineering of reward functions prevents us from end-to-end learning from pixels, if the reward function itself requires a dedicated perception pipeline.  To address these problems, we propose an unsupervised skill learning method that 
aims to discover and learn transferable skills by watching videos. The learned embedding is then used to guide an RL-agent in order to solve a wide range of tasks by composing previously seen skills.

Prior work in visual representation learning for deriving reward functions relied on self-supervised objectives~\cite{Sermanet2017TCN,watter2015embed,finn2017deep,Yu2019, Sermanet2017Rewards} and focused on single tasks. Not only is this inefficient, but also limits the versatility and adaptivity of the systems that can be built.
Thus, we consider the problem of learning a multi-skill embedding without human supervision.

In this paper, we present a novel approach called Adversarial Skill Networks (ASN). In order to learn a task-agnostic skill embedding space, our method solely relies on unlabeled multi-view observations. Hence, it does not require correspondences between frames and task IDs nor any additional form of supervision or instrumentation. We combine a metric learning loss, which utilizes temporal video coherence, with an entropy-regularized adversarial skill-transfer loss.  Our results indicate that the learned embedding can be used not only to train RL agents for tasks seen during the training of the embedding, but also for novel tasks that require a composition of previously seen skills. 

\begin{figure}[t]
   \includegraphics[width=\linewidth]{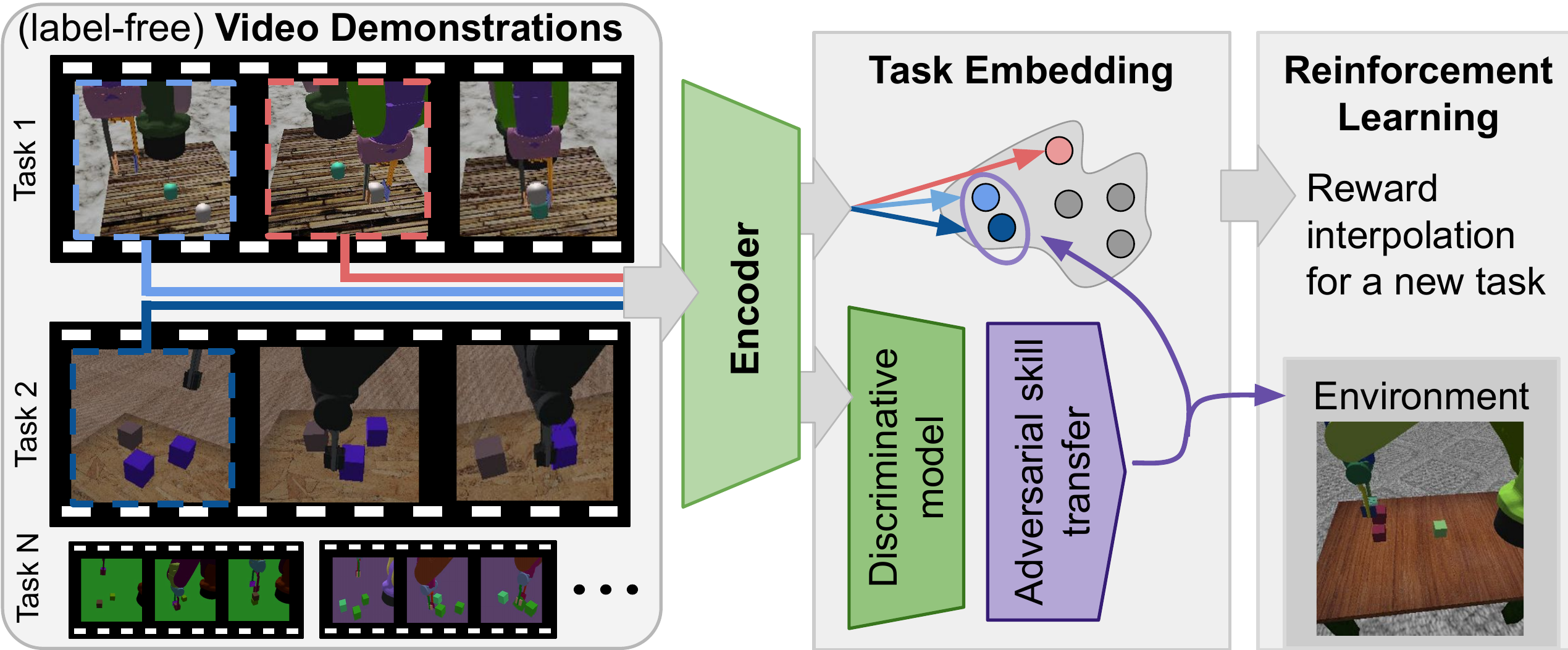}
   \caption{Given the demonstration of a new task as input, Adversarial Skill Networks yield a distance measure in skill-embedding space which can be used as the reward signal for a reinforcement learning agent for multiple tasks.}
   \label{fig:overview}
\end{figure}

In extensive experiments, we demonstrate both qualitatively and quantitatively that our method learns transferable skill embeddings for simulated and real demonstrations without the requirement of labels. We represent the skill embedding as a latent variable and apply an adversarial entropy regularization technique to ensure that the learned skills are task independent and versatile and that the embedding space is well formed. We show that the learned embedding enables training of continuous control policies with PPO~\cite{ppo} to solve novel tasks that require the interpolation of previously seen skills. Training an RL-agent to re-use skills in an unseen task, by using the learned embedding space as a reward function, solely requires a single video demonstrating the novel task. This makes our method readily applicable in a variety of robotics scenarios.

\section{Related Work}

Our work is primarily concerned with learning representations that enable a robot to solve multiple tasks by re-using skills without human supervision, thus falling under the category of self-supervised robot learning~\cite{Sermanet2017TCN,finn2017deep,Sermanet2017Rewards, mees19iros}. There exists a large body of work for learning representations through autoencoders~\cite{finn2016deep, burgess2019monet}, pre-trained supervised features~\cite{Sermanet2017Rewards}, spatial structure~\cite{finn2016deep, jonschkowski2017pves} and state estimation from vision~\cite{opean-ai-hand}. Compared to these  approaches, we take multiple tasks into account to learn a skill embedding before training a reinforcement learning agent with a self-supervised vision-based training signal.

Further approaches attempt to derive data-driven reward functions~\cite{tdc,singh2019, Yu2019, Sermanet2017Rewards, finn2017deep}  by providing a label-free training signal from video or images to minimize human supervision. Related to our work, Sermanet \emph{et al.}~\cite{Sermanet2017Rewards} provide  reward functions by identifying key intermediate steps for one task from multiple video examples. Other methods~\cite{singh2019, Yu2019, watter2015embed,finn2016deep} use images of goal examples to construct a task objective for goal reaching tasks such as pushing. Atari video games are solved in \cite{ibarz2018reward,tdc} by constructing an objective from human demonstration videos. However, it is unclear whether the reward signal reflects a good performance when transferring it to a real-world robotic task.
Our model is able to find task specific features and generalizes to unseen objects, viewpoints and backgrounds.

Existing methods for learning reusable skill embeddings make use of entropy-maximization of the policy~\cite{hausman2018learning, haarnoja2018composable, esteban2019hierarchical} and therefore allow for policy interpolation. Haarnoja \emph{et al.}~\cite{haarnoja2018composable} use a composition of soft Q-functions to create a policy that reaches a new goal. 
Hausmann \emph{et al.}~\cite{hausman2018learning} propose a hierarchical reinforcement learning approach that utilizes two embedding networks and an entropy regularization on the policy to cover a latent space with different skill clusters. Orthogonal to our work, these methods rely on previously designed reward functions. 

Most related to our approach is the work by Sermanet \emph{et al.}~\cite{Sermanet2017TCN} that introduces \textit{Time-Constrative Networks}  (TCN) and a triplet loss  combined with multi-view metric learning to increase the distance of embeddings for transitions far apart in time. The learned metric can then be used as a reward signal within a RL-setup by minimizing the distance to a visual demonstration. However, TCN focuses only on the single-task setting and does not leverage information from previously learned skills.
Dwibedi \emph{et al.}~\cite{dwibedi2018learning} extend TCN using multiple frames (mfTCN). In contrast to our approach, the embedding is not used as a (label-free) reward signal.

In addition to the metric loss, we use an adversarial loss term~\cite{gan,catgan15,revgrad,revgradcond,domainfake} as a regularization technique. The adversarial loss was introduced for \gls{gan}~\cite{gan} and domain adaptation~\cite{revgrad,revgradcond,domainfake}. Similar to the problem of domain adaptation we have multiple videos for different task domains. For domain adaptation, multiple approaches~\cite{revgrad,revgradcond} use a gradient reversal layer and  Tzeng \emph{et al.}~\cite{domainfake} exploit a \gls{gan}-based loss. Springenberg \emph{et al.}~\cite{catgan15} introduce an objective function for label free classification by extending GANs to categorical distributions. Our approach uses a similar adversarial loss to learn a reusable skill embedding for different task domains.

In contrast to these previously described approaches, we propose a method to learn skills from video by a composition of metric learning and an entropy-regularized adversarial skill-transfer loss. Our method not only allows for the representation of multiple task-specific reward functions, but also builds upon this information in order to interpolate between learned skills, see \Cref{fig:overview}.

\section{Learning a Transferable Skill Embedding}
\begin{figure*}[ht]
    \centering
\begin{AnnotatedImage}[width=0.88]{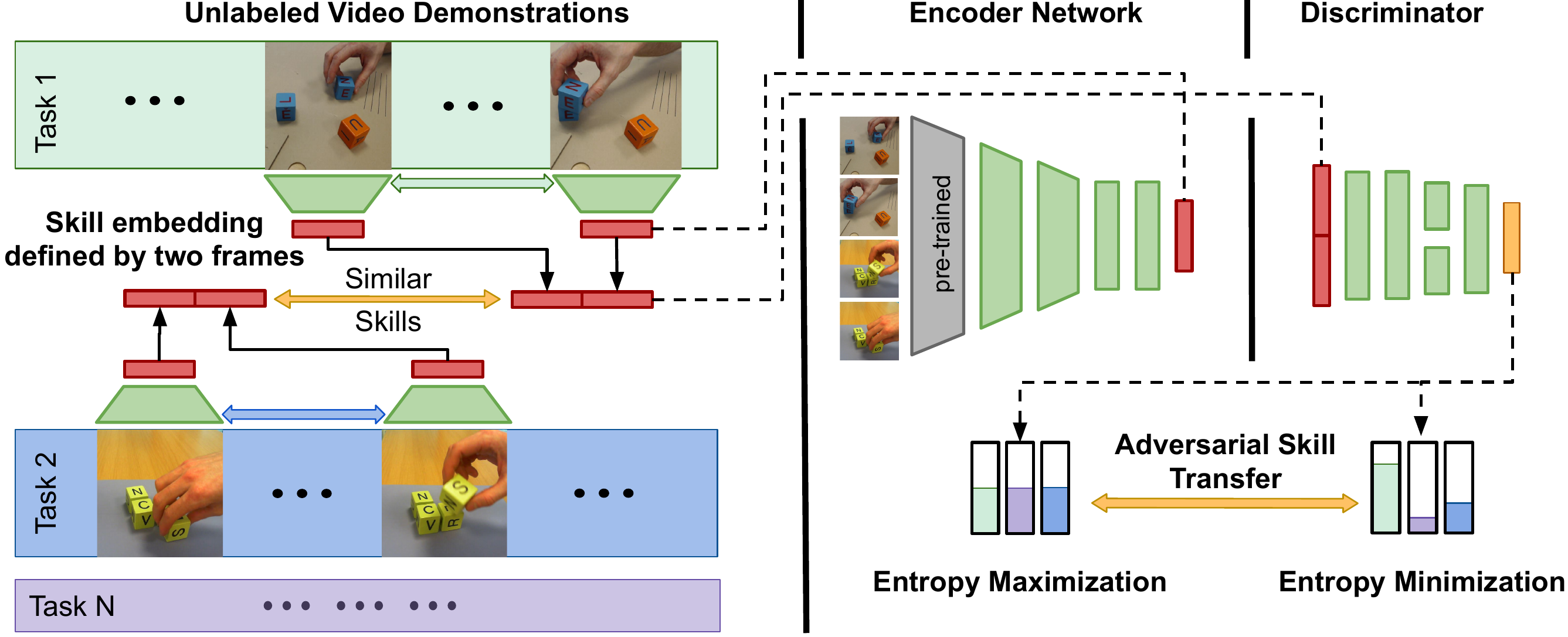}
\annotate (G) at (0.3,0.67){\small$\Delta t$};
\annotate (G) at (0.2,0.39){\small$\Delta t$};
\annotate (G) at (0.03,0.97){\small$t=0$};
\annotate (G) at (0.03,0.36){\small$t=0$};
\end{AnnotatedImage}
   \caption{Structure of Adversarial Skill Networks: We learn a skill metric space in an adversarial framework. The encoding part of the network tries to maximize the entropy to enforce generality. The discriminator, which is not used at test time, tries to minimize the entropy of its prediction to improve recognition of the skills. Finally, maximizing the marginal class entropy over all skills leads to uniform usage of all task classes. Please note that no information about the relation between frames and the tasks they originated from is needed.}
   \label{fig:architecture}
\end{figure*}
\label{sec:result}

The main incentive of our method is a more general representation of skills that can be re-used and applied to novel tasks. In this work, we define tasks to be composed of a collection of skills. Since we approach this problem in an unsupervised fashion, we do not need any labels describing relations between different task videos or even for different examples of the same task.
In our approach, we are interested in the following properties for a learned embedding space:

\noindent\textbf{i) versatility:} Multiple tasks can be represented in the same embedding space.

\noindent\textbf{ii) skill representation:} A skill can be described by the embedding of two sequential frames, with a  time delay in-between (stride).

\noindent\textbf{iii) generality:} The learned embedding space should generalize to unseen objects, backgrounds and viewpoints.

\noindent\textbf{iv) task independent skills:} It should not be possible to distinguish similar skills from different task domains, i.e., the same skill executed in different environments should ideally have an identical embedding.

\subsection{Adversarial Skill Networks}
We propose \gls{asn} to achieve a novel skill representation, which takes these properties into account. We combine a metric learning loss, which utilizes temporal video coherence to learn a state representation, with an entropy-regularized adversarial skill-transfer loss. An overview can be seen in \Cref{fig:architecture}.

To transfer similar skills without any label information to unseen tasks, one needs to learn generalized skills that should neither be task- nor domain-specific, following property \textbf{iv)}.  Since the true class distribution over skills is not known, this problem can naturally be considered as a ``soft'' probabilistic cluster assignment task.

To solve this, our method introduces a novel entropy regularization by jointly training two networks in an adversarial manner: an encoder network $E$ and a discriminator $D$. Given two sequential frames $(v,w)$, which are separated by a temporal stride $\Delta t$, we define an unlabeled skill embedding $\mathbf{x}=(E(v), E(w))$ and collection of skills $\mathcal{X} = \{\mathbf{x}^1,...,\mathbf{x}^N\}$ representing the different tasks. The encoder network  embeds single frames of the dimension $d_1\times d_2$ into a lower-dimensional representation of size $n$, i.e. $E$:  $\mathbb{R}^{d_1\times d_2} \to \mathbb{R}^n$. 
We compute Euclidean distances in the embedding space to compare the similarity of frames. The discriminator network takes two concatenated embedded frames that define an unlabeled skill $\mathbf{x}$ as  input and outputs $y_c$, the probability of the skill being originated from task $c$. Formally, we require $D(\mathbf{x}) \in \mathbb{R}^C$ to give rise to a conditional distribution over tasks $\sum_{c=1}^{C} p (y_c =c \mid \mathbf{x}, D) =  1$. Although we define this hyper-parameter a priori as the number of tasks contained in a training set, we observed minor performance drops setting it to a value with small deviation from the true number of tasks. Most importantly, \gls{asn} does not need a task label for the demonstration videos.

The encoder parameters are updated using a metric learning loss and maximization of the entropy of the discriminator output. In order to capture the temporal task information, we use a modified version of the lifted structure loss \cite{DBLP:journals/corr/SongXJS15}. Given two view-pairs $(v_1, v_2)$, synchronized videos from different perspectives, we attract frames that represent the same temporal task state and repulse temporal neighbors, given a constant margin $\lambda$, i.e.
\begin{dmath}
    \mathcal{L}_{\text{lifted asn}}= \sum_{i=1}^M\left(\log\sum\limits_{y_k=y_i} \Biggl(\mathrm{e}^{\lambda-S_{ik}}+\mathds{1}_{S_{ik}>\xi}\cdot S_{ik}\Biggr)+\log\sum\limits_{y_k\neq y_i}\mathrm{e}^{S_{ik}}\right),
\end{dmath}
for $M$ frames ($x_1,x_2...x_M$) and $S_{ij} = E(x_i) \cdot E(x_j)$,  as a dense squared pairwise similarity distance matrix of the batch and $\xi$ a similarity threshold. Additionally, we introduce a constraint that bounds the distance between two positive view-pairs. This constraint is tailored to account for high variance in the learned distance metric. By penalizing large distances of positive view-pairs, we aim at smoother transitions between similar states in a RL setting.

The discriminator network minimizes the entropy given an unlabeled skill embedding $\mathbf{x}$ to be certain about which task $C$ the skill originated from. Note that the discriminator $D$ is utilized only during training. Without any additional label information about the $C$ classes, we cannot directly specify which class probability $p(y_c=c \mid \mathbf{x}^i, D)$ should be maximized for any given skill $\mathbf{x}$. We make use of information theoretic measures on the predicted class distribution to group the unlabeled skills into well separated categories in the skill embedding space without explicitly modeling $p(\mathbf{x})$.  Specifically, if we want the discriminator to be certain for the class distribution $p(y_c=c\mid \mathbf{x}^i, D)$, this corresponds to minimizing the Shannon information entropy $H[p(y_c\mid \mathbf{x},D)]$, as any draw from this distribution should most of the times result in the same class. On the other hand, if we want the encoder to learn generalized skill representations to meet requirement \textbf{iv)}, it should be uncertain of how to classify the unlabeled skills. Thus, the encoder tries to maximize the entropy $H[p(y_c\mid \mathbf{x},D)]$, which at the optimum will result in a uniform conditional distribution over task classes. Concretely, we define the empirical estimate of the conditional entropy over embedded skill examples $\mathcal{X}$ as: 
\begin{equation}
    \begin{aligned}
\mathbb{E}_{\mathbf{x}\sim\mathcal{X}} \Big[H[p(y_c\mid \mathbf{x},D)] \Big] =  \frac{1}{N}\sum_{i=1}^N H\left[ p(y_c\mid \mathbf{x}^i, D)\right] \\
= \frac{1}{N}\sum_{i=1}^N - \sum_{c=1}^C p(y_c=c\mid \mathbf{x}^i, D) \log p(y_c=c\mid \mathbf{x}^i, D).
    \end{aligned}
\end{equation}
With an additional regularizer we enhance the equal usage of all task classes, corresponding to maximizing a
uniform marginal distribution:
\begin{equation}
H_{\mathbf{x}}[p(y_c\mid D)]= H\left[\frac{1}{M}\sum_{i=1}^M p(y_c\mid \mathbf{x}^i, D)\right],\\
\end{equation}
where $M$ is set to the number of independently drawn samples~\cite{catgan15}.
 
In order to disentangle the learned metric and the mapping to task IDs, we add a sampled latent variable  $z=\mu + \sigma \odot \mathcal{E}$ and  $\mathcal{E}\sim N(0,1)$, where $D$ estimates $\mu, \sigma$ of a Gaussian distribution.
We use the re-parameterization trick to back-propagate through the random node~\cite{vae1}.
With the Kullback-Leibler divergence regularization for $z$ we force $D$ to find similar properties describing the skills.
Without this objective, similar skills could end up represented  far away from each other in the skill embedding space.

This leads to the following objectives for the encoder $E$  and discriminator $D$: 
\begin{equation}
    \begin{aligned}
\mathcal{L}_{KL}={}&D_{KL}[p(z\mid \mathbf{x})||p(z)],\\
\mathcal{L}_D = {}& -H_{\mathbf{x}}\Big[p(y_c\mid D)\Big] + \mathbb{E}_{\mathbf{x}\sim\mathcal{X}}\Big[H[ p(y_c\mid \mathbf{x},D)]\Big] \\ &\qquad +\beta\mathcal{L}_{KL} \text{ and} \\
\mathcal{L}_E = {}& H_{\mathbf{x}}\Big[p(y_c\mid D)\Big] + \mathbb{E}_{\mathbf{x}\sim\mathcal{X}} \Big[H[p(y_c\mid \mathbf{x},D)]\Big] \\
& \qquad -\alpha\mathcal{L}_{\text{lifted asn}}. \\
    \end{aligned}
\end{equation}

We therefore optimize the discriminator and the encoder according to:
\begin{equation}
\min_D\mathcal{L}_D 
\end{equation}
and
\begin{equation}
\max_E \mathcal{L}_E.
\end{equation}

\subsection{Implementation Details}

The encoder network is inspired by \gls{tcn}~
\cite{Sermanet2017TCN}. We use an Inception network as a feature extractor~\cite{inception},
which is initialized  with ImageNet pre-trained weights.
The feature extractor is followed by two convolutional layers and a spatial softmax layer for dimension reduction.
Finally, after a \gls{fc} layer, the model outputs the embedding vector for a frame. 
For all experiments, we use $\alpha=0.1$, $\beta=1.0$, $\lambda=1.0$ and an embedding size of 32.
The discriminator consists of two \gls{fc} layers to estimate $\mu$ and $\sigma$ of a Gaussian 
distribution, followed by two layers to output the task ID. 
We use dropout for regularization. 

We train the encoder and discriminator networks with the Adam optimizer and a
 learning rate of $0.001$. 
A training batch contains 32 frames from $n=4$ different view pairs.
We load real-world data from video files and sample the simulated data from uncompressed image files.
Training directly on images, ensures that our model is not learning any bias introduced by video compression techniques.
For frames from the training set, we randomly change brightness, contrast and saturation and randomly mirror frames horizontally. 
For real-world data, additional training frames are cropped randomly. 
We train on images of the size $299\times 299\times 3$ pixels.
For simulated data the discriminator network is only updated with successful task demonstrations, 
since only they contain the skills we want to transfer.
After data augmentation, the frames of a batch are normalized on each RGB channel using the $\mu$ and $\sigma$ 
of the ImageNet dataset.%


\section{Experimental Results}

We evaluate the performance of our \gls{asn} model on two data sets, see Figure \ref{fig:dataset}.
\newlength{\tasksubfigwidth}
\setlength{\tasksubfigwidth}{0.495\textwidth}
\begin{figure}[h]
    \centering
    \begin{subfigure}[b]{\tasksubfigwidth}
    \centering
      \includegraphics[width=0.7\linewidth]{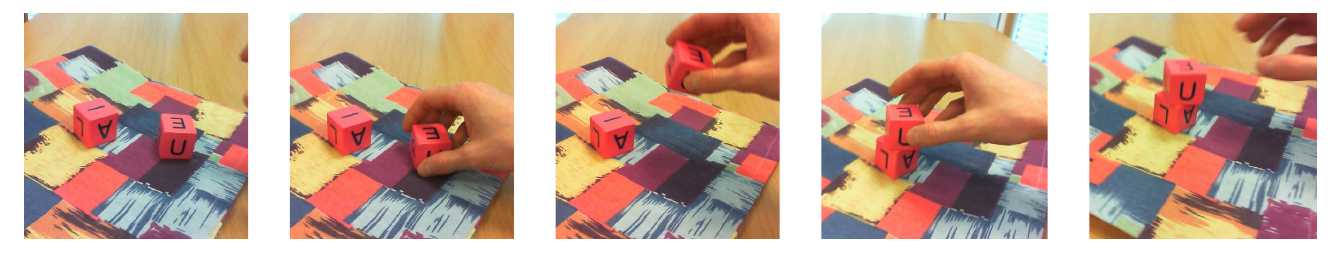}
      \includegraphics[width=0.7\linewidth]{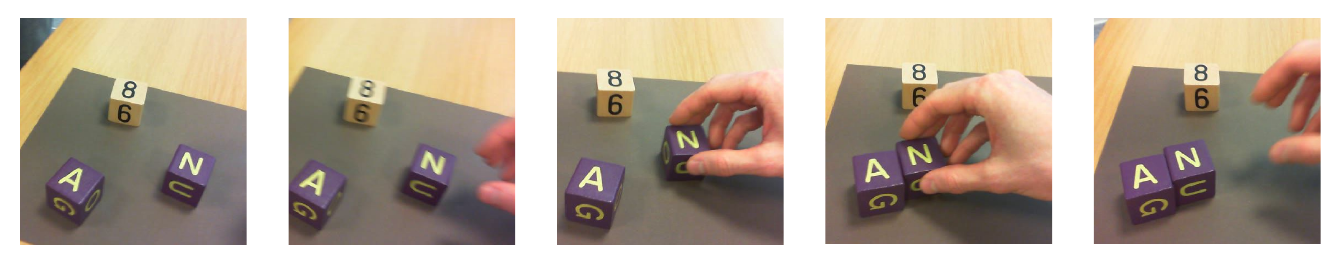}
      \includegraphics[width=0.7\linewidth]{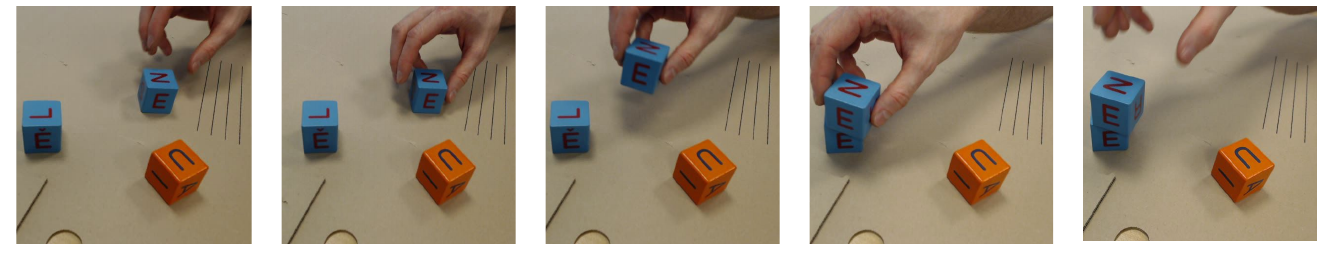}
      \includegraphics[width=0.7\linewidth]{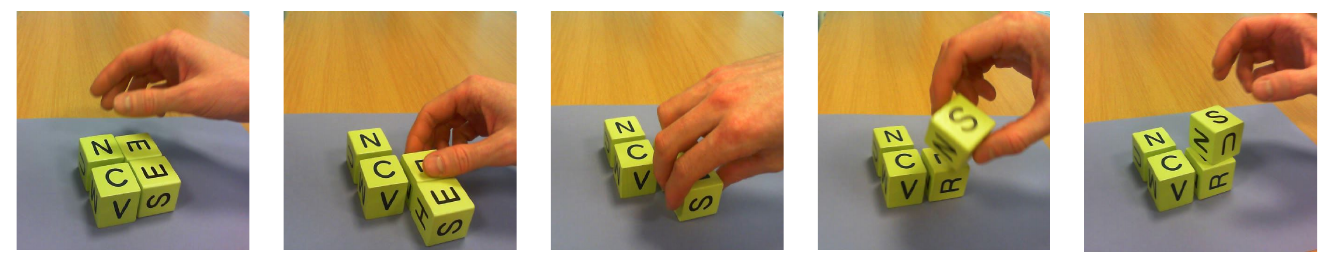}
      \caption{Real block tasks}\label{fig:Ng1}
    \end{subfigure}
    \begin{subfigure}[b]{\tasksubfigwidth}  
    \centering
      \includegraphics[width=0.7\linewidth]{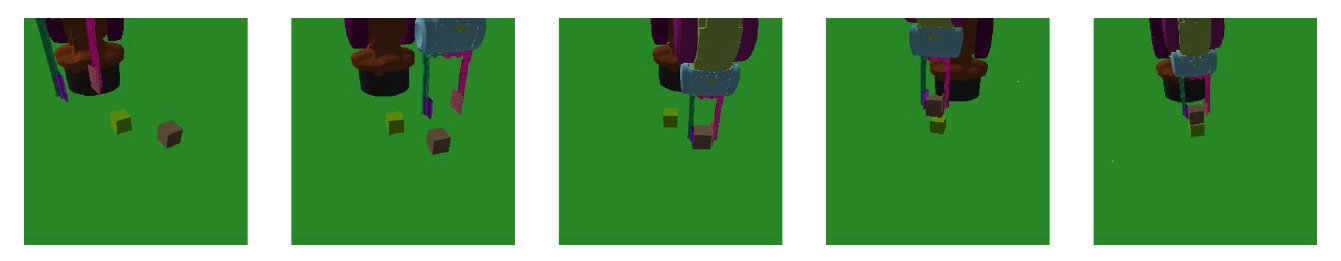}
      \includegraphics[width=0.7\linewidth]{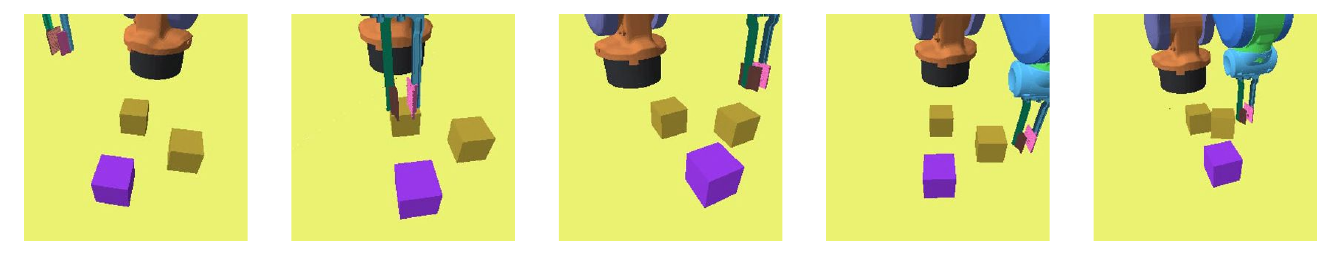}
      \includegraphics[width=0.7\linewidth]{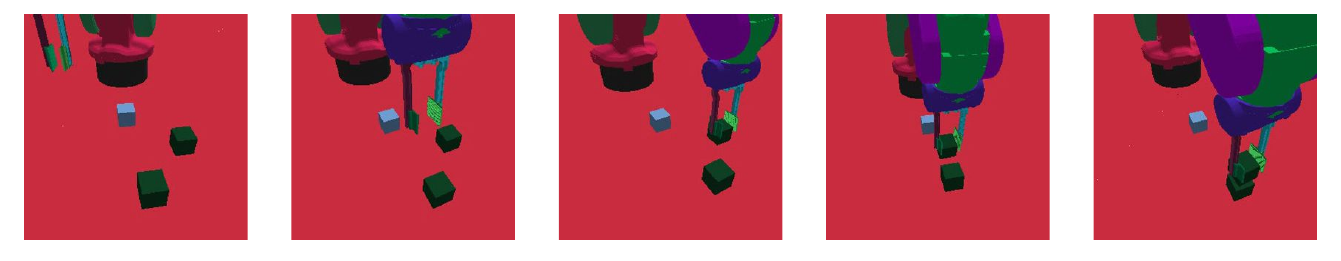}
      \includegraphics[width=0.7\linewidth]{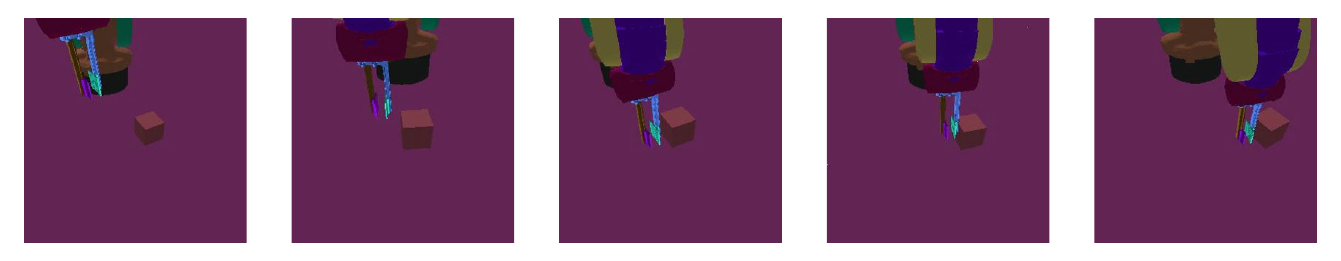}
      \caption{Simulated block tasks}\label{fig:Ng2}
    \end{subfigure}
    \caption{Visualization of the multi-task datasets used in this work.}\label{fig:dataset}
\end{figure}

     The first data set consists of three simulated robot tasks: stacking (A), color pushing (B) and color stacking (C). The data set contains 300 multi-view demonstration videos per task. The tasks are simulated with PyBullet. Of these 300 demonstrations, 150 represent unsuccessful executions of the different tasks. We found it helpful to add unsuccessful demonstrations in the training of the embedding to enable training RL agents on it. Without fake examples, the distances in the embedding space for states not seen during training might be noisy. In the initial phase of training, however, the policy to be learned mostly visits areas of the state-action space which are not covered by the (successful) demonstration. Hence, it is important to have unsuccessful examples in the training set. The test set contains the manipulation of blocks. Within the validation set, the blocks are replaced by cylinders of different colors.
     
     The second data set includes real-world human executions of the simulated robot tasks (A, B and C), as well as demonstrations for a task where one has to first separate blocks in order to stack them (D). Each task contains 60 multi-view demonstration videos, corresponding to 24 minutes of interaction. The test set contains blocks of unseen sizes and textures, as well as unknown backgrounds, in order to evaluate the generality of our approach. 

\begin{figure*}[t]
\centering
\setlength{\tabcolsep}{10pt}
\begin{tabular}{cccc}
\begin{overpic}[width=0.18\linewidth]{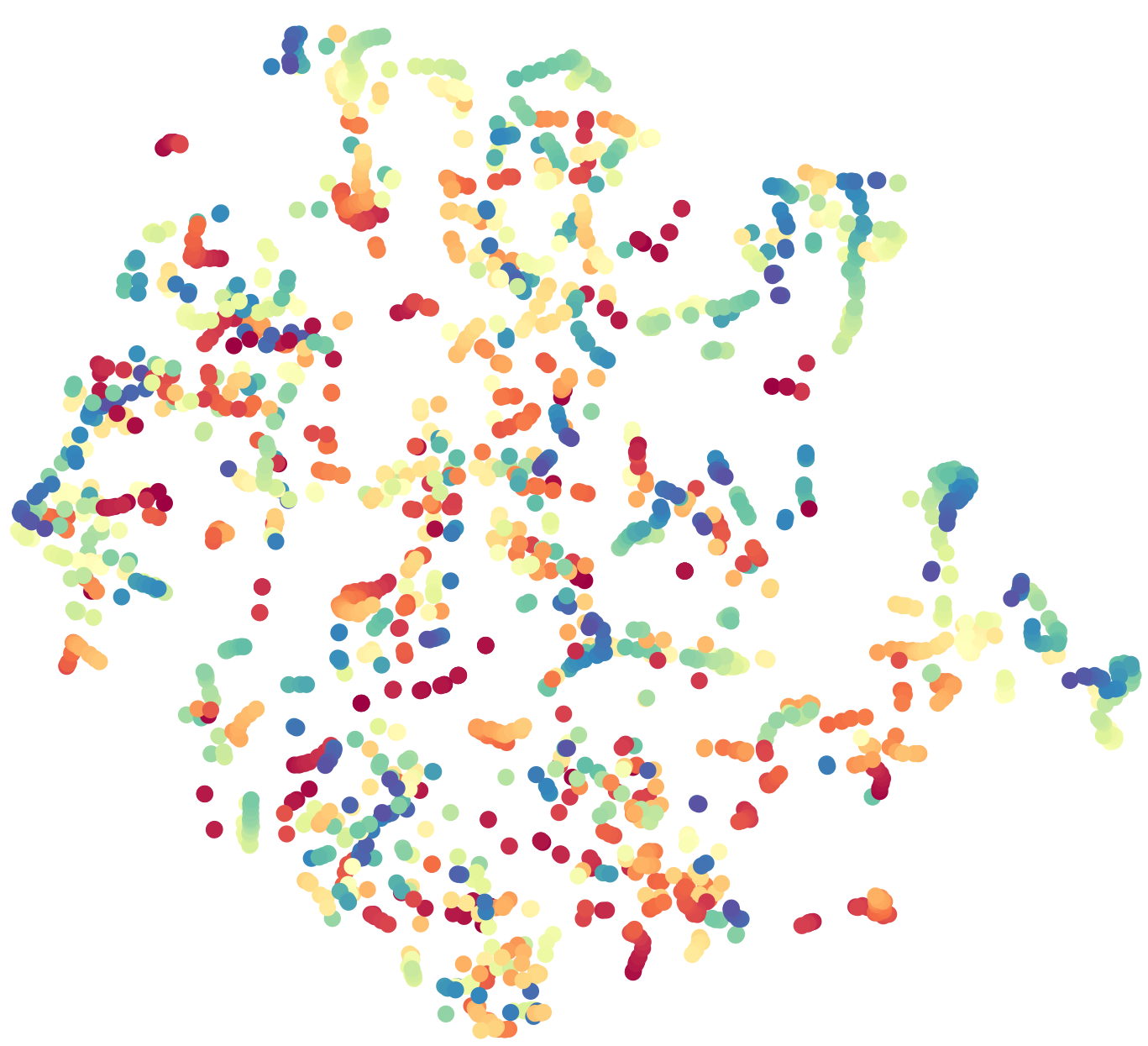}
\put (10,95) {ImageNet weights}
\end{overpic}&
\begin{overpic}[width=0.18\linewidth]{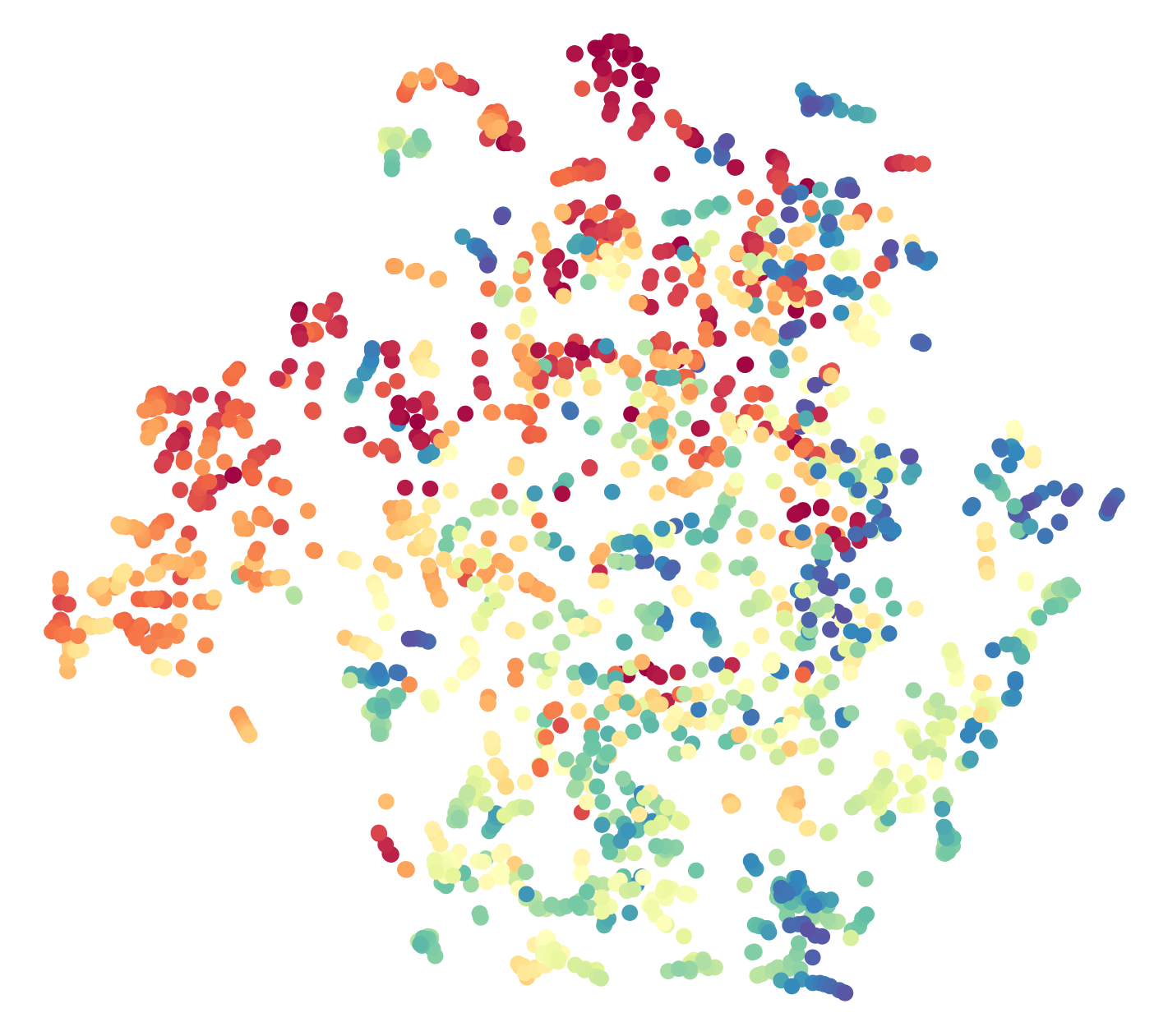}
\put (40,95) {TCN}
\end{overpic}&
\begin{overpic}[width=0.18\linewidth]{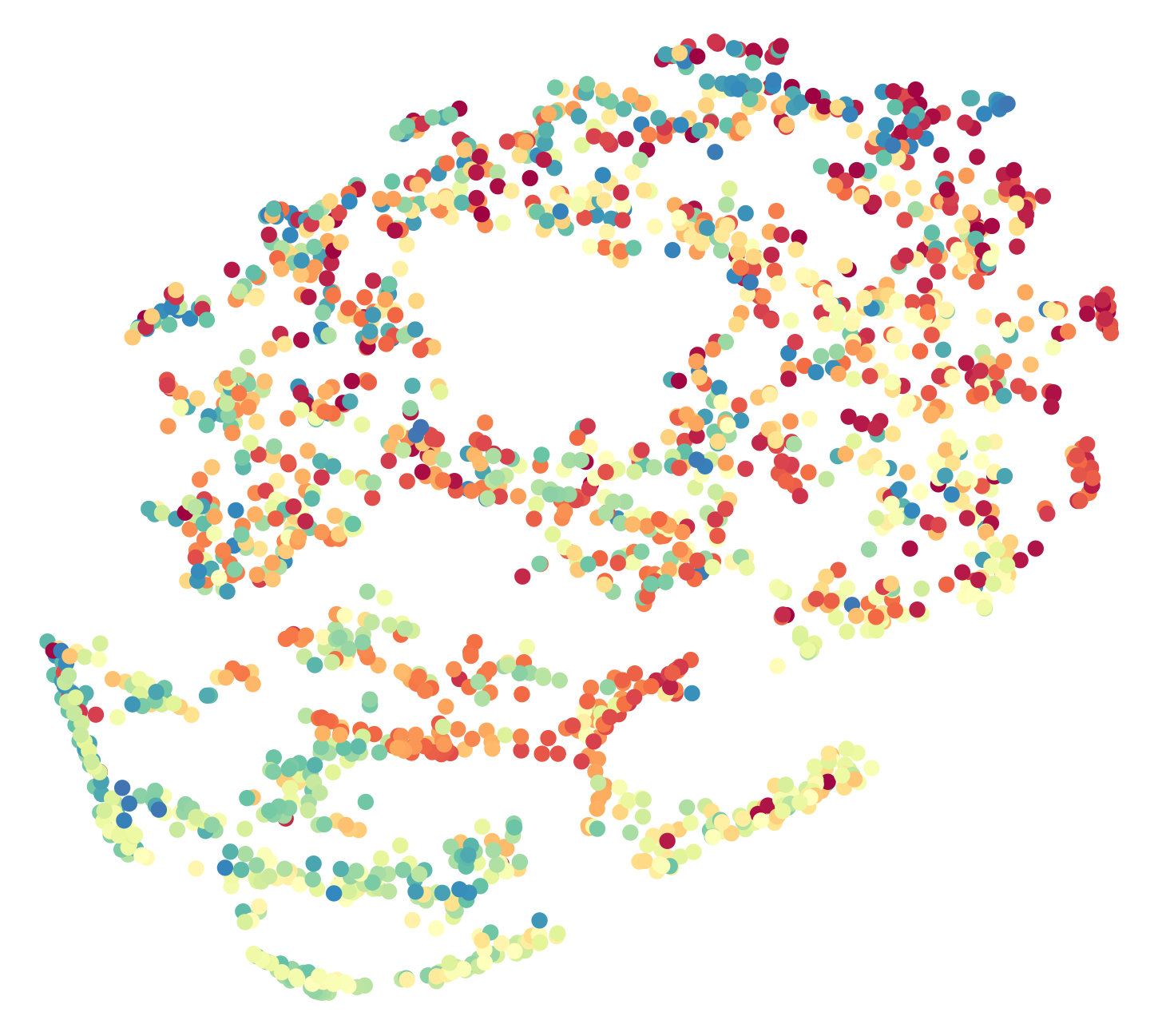}
\put (40,95) {mfTCN}
\end{overpic}&
\begin{overpic}[width=0.18\linewidth]{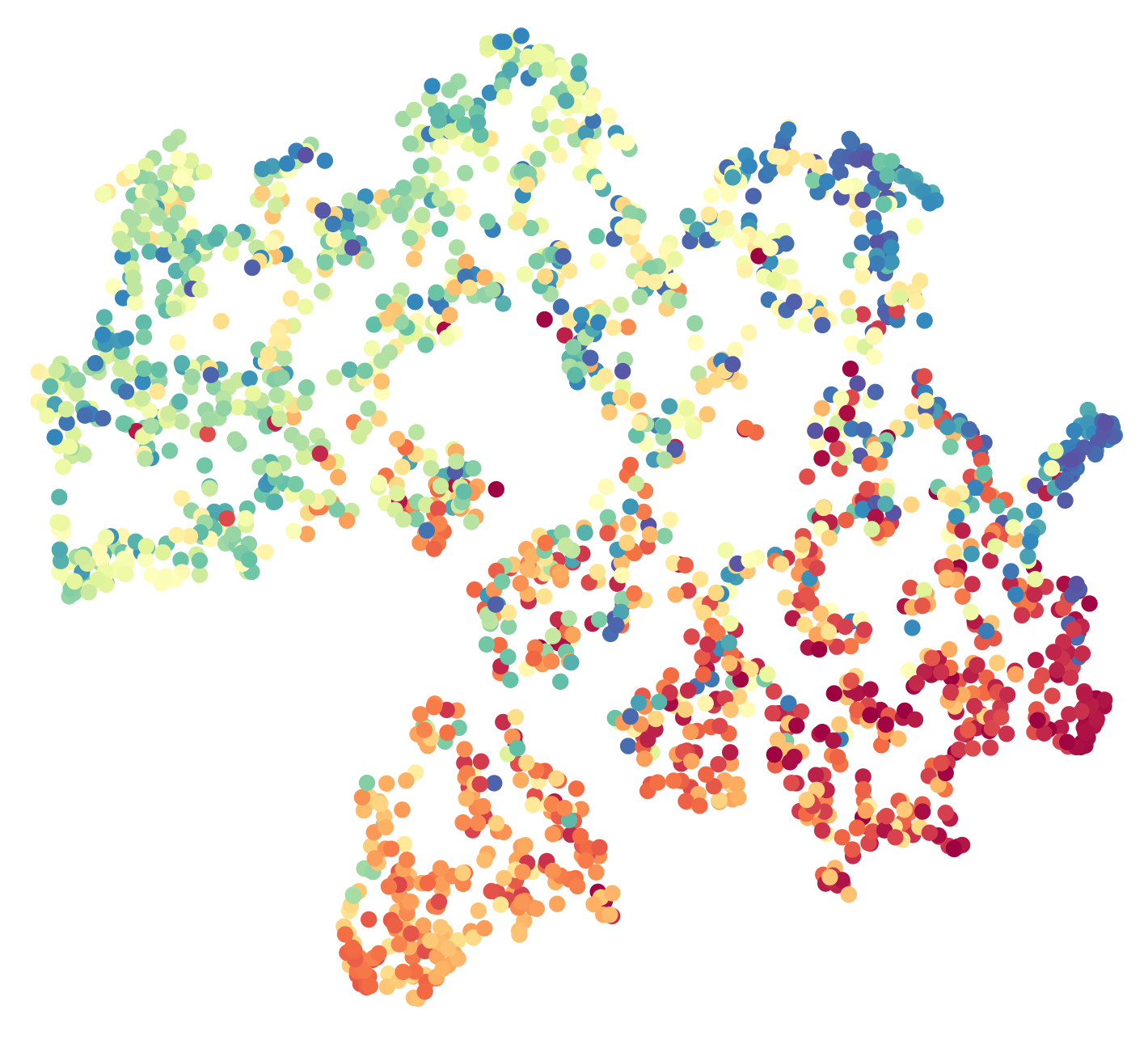}
\put (40,95) {ASN}
\end{overpic}\\
\end{tabular}
\begin{tabular}{c}
\begin{overpic}[width=0.9\linewidth, height=0.6cm]{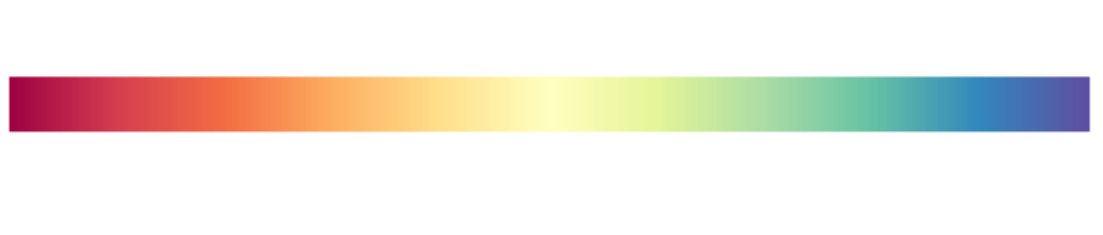}
\put (1,-1) {$t=0$}
\put (96,-1) {$T$}
\end{overpic}
\end{tabular}
\caption{t-SNE of an \textbf{unseen} color stacking video for models trained on block stacking and color pushing. Our ASN model maintains the temporal coherence of the task better than the baselines. The colorbar indicates the temporal task progress.}
\label{fig:tsne}
\end{figure*}

\subsection{Quantitative Evaluation}
We measure the performance of our models based on the alignment loss, following Sermanet \emph{et al.}~\cite{Sermanet2017TCN},  to determine how well two views of a video are aligned in time.  We take advantage of the
fact that frames in both videos are synchronized with each other to get alignment labels for free. We try to sequentially align two view pairs by finding the nearest neighbor in the embedding space normalizing the distances by the demonstration length. After embedding the two videos in our skill embedding space, we search for each frame $t^i_j$ in the first video $i$ the nearest neighbour in the second view and retrieve its time index $t^{nn}_j$. Thus, for video $i$ the aligment loss is defined by:
 \begin{equation}
     \text{align}_i = \frac{\sum_{j=1}^F| t^i_j - t^{nn}_j|}{F},
 \end{equation}
for a video of length $F$. As Sermanet \emph{et al.}~\cite{Sermanet2017TCN} have shown, the alignment loss reflects the quality of the reward signal within a RL setup.  Instead of evaluating the alignment loss on the same task as the embedding was trained on, we measure how well view pairs of novel, unseen tasks are aligned. This form of zero-shot evaluation is very challenging, as it requires the combination of previously seen skills.  Adversarial Skill Networks yield the best performance for transfer in the simulated robot setting, which can be seen in \Cref{tab:sim}. Please note the lower bound (0.081) of alignment loss for single-task TCN. In contrast to TCN trained on multiple tasks, our model gets very close (0.099) despite not being trained on the task. Furthermore, our approach outperforms TCN in both the real robot multi-task setup, as well as in the transfer task, which is depicted in \Cref{tab:real}. We also compare against the different metric learning losses and show that the lifted loss in combination with the bound for positive view-pairs outperforms other methods.
 \begin{table}[h]
  \centering
    \begin{tabular}{l*{3}{c}}
      \toprule
         \multirow{2}{*}{\textbf{Model}}  & \multicolumn{3}{c}{\textbf{Task Combination, Train$\,\to\,$Test }}  \\
                                          & C$\,\to\,$C & A,B,C$\,\to\,$A,B,C & A,B$\,\to\,$C     \\
        \midrule
         Inception-ImageNet~\cite{inception} & 0.29 & 0.31 & 0.29  \\
        \acrshort{tcn} - lifted~\cite{Sermanet2017TCN} & 0.081 & 0.058 & 0.112 \\
        \acrshort{asn}  & -& 0.056 & \textbf{0.099}  \\
        \bottomrule
    \end{tabular}
    \caption{Test alignment loss for the  simulated robot multi-task dataset, which includes fake examples,
             Tasks: A: 2 block stack, B: 3 block color push sort, C: 3 block color stack sort.}
    \label{tab:sim}   
\end{table}

A visualization of the learned embedding space is  depicted in \Cref{fig:tsne}. \gls{asn} can represent the temporal relations of an unseen task better than TCN or multi-frame TCN, leading to more meaningful distance measures in embedding space. Our proposed multi-task setup is able to reflect skills needed to solve the unseen task and thus can generalize better, while maintaining the temporal coherence of the unseen task.
\begin{table}[h]
  \centering
    \begin{tabular}{l*{3}{c}}
      \toprule
            \multirow{2}{*}{\textbf{Model}} & \multicolumn{3}{c}{\textbf{Task Combination, Train$\,\to\,$Test}} \\
                       & A,B,C$\,\to\,$C & A,B$\,\to\,$C  & A,B,D$\,\to\,$C    \\
        \midrule
        \acrshort{tcn} - triplet~\cite{Sermanet2017TCN} & 0.186 & 0.21 & 0.218 \\
        \acrshort{tcn} - lifted~\cite{Sermanet2017TCN} & 0.171 & 0.20 & 0.187 \\       
        \acrshort{tcn} - npair~\cite{Sermanet2017TCN} & 0.221 & 0.209 & 0.221 \\
        \acrshort{rnntcn} - lifted~\cite{dwibedi2018learning}  & 0.174 & 0.23 & 0.22 \\
        \acrshort{asn} - normal lifted &  0.168 &  0.183 & 0.181\\
        \acrshort{asn}  &  \textbf{0.150} & \textbf{0.180} & \textbf{0.165}\\
        \bottomrule
    \end{tabular}
    \caption{Test alignment loss real-world block tasks,
             Tasks: A: 2 block stack, B: 3 block color push sort, C: 3 block color stack sort,
              D: 4 block separate to stack.}
    \label{tab:real}   
\end{table}
\subsection{Ablation studies}
 To analyze the influence of our different building blocks on the learned embedding, we conducted several experiments, see  \Cref{tab:ablation}. Our results indicate that it is of benefit to describe a skill with a growing stride, so as to cover macro-actions describing events longer in time. In order to keep the embeddings of these skill frames of higher stride aligned, the KL-divergence is shown to be an effective regularization technique, yielding the lowest alignment loss. Furthermore, it seems to be enough to describe a skill by only the start and end frames. A single frame seems to provide too little information whereas using four frames proves to make the state space too high dimensional. 
\begin{table}[h]
    \centering
    \setlength{\tabcolsep}{3.pt}
 \begin{tabular}{l*{4}{c}}
      \toprule
            \multirow{2}{*}{\textbf{Regularization}} & \multirow{2}{*}{\textbf{\#Domain frames}}  &\multirow{2}{*}{\textbf{Stride}}  & \multicolumn{1}{c}{\textbf{Real block tasks}}\\
                       & & &  A,B,D$\,\to\,$C \\
        \midrule
        KL  & 1 & - & 0.187 \\
        KL  & 4 & 5 & 0.185 \\  
        KL  & 2 & 5 & 0.168 \\
        KL  & 2 & 15 & \textbf{0.165}  \\
        FC  & 2 & 15 & 0.1987  \\
        KL w/o encoder  entropy & 2 & 15 & 0.186  \\
        KL w/o   entropy & 2 & 15 & 0.177  \\
        \bottomrule
    \end{tabular}
    \caption{Ablation studies: transfer loss for different regularization techniques and skill definitions.}
    \label{tab:ablation}
\end{table}
\begin{figure*}[t]
    \centering
       \includegraphics[width=0.82\linewidth,trim=0 0 0 22, clip]{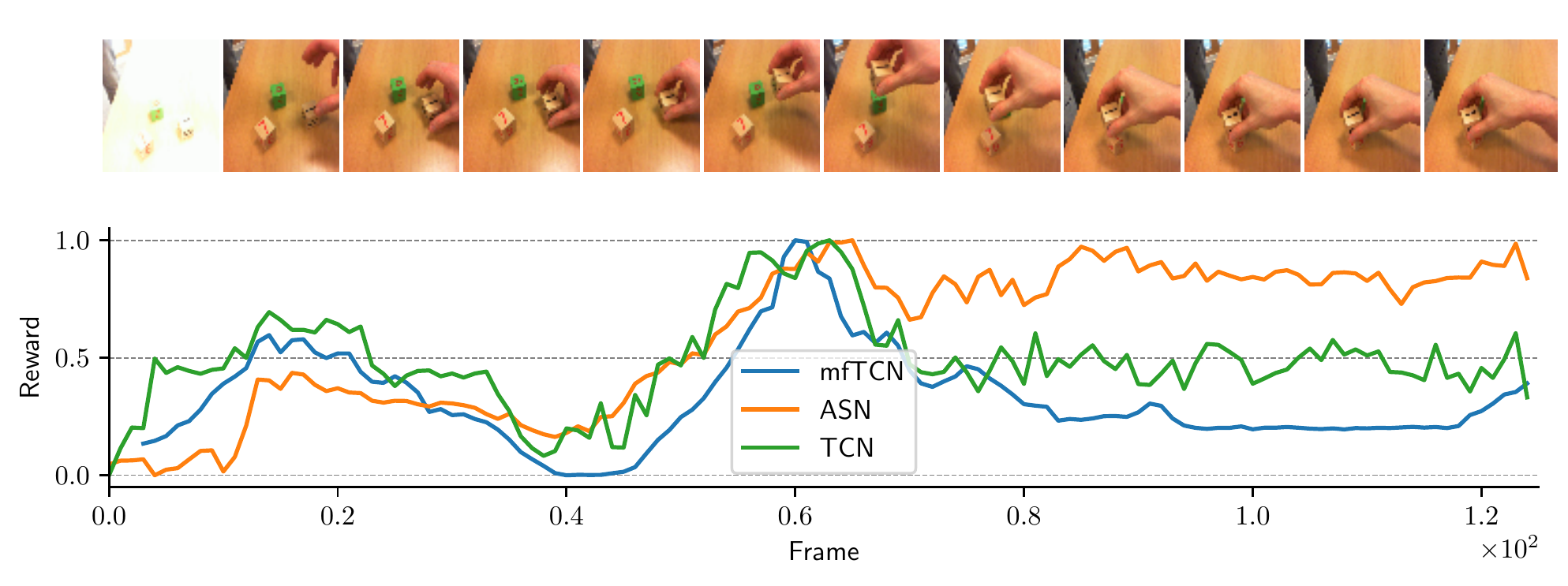}
    \caption{Reward plot for a novel color stacking task C for models trained on tasks A, B and D. The reward is based on the distance to a single goal frame from a different perspective.}
    \label{fig:rw}
\end{figure*}
\subsection{Learning Control Policies}
\begin{figure}[h]
    \centering
    \includegraphics[width=0.49\linewidth]{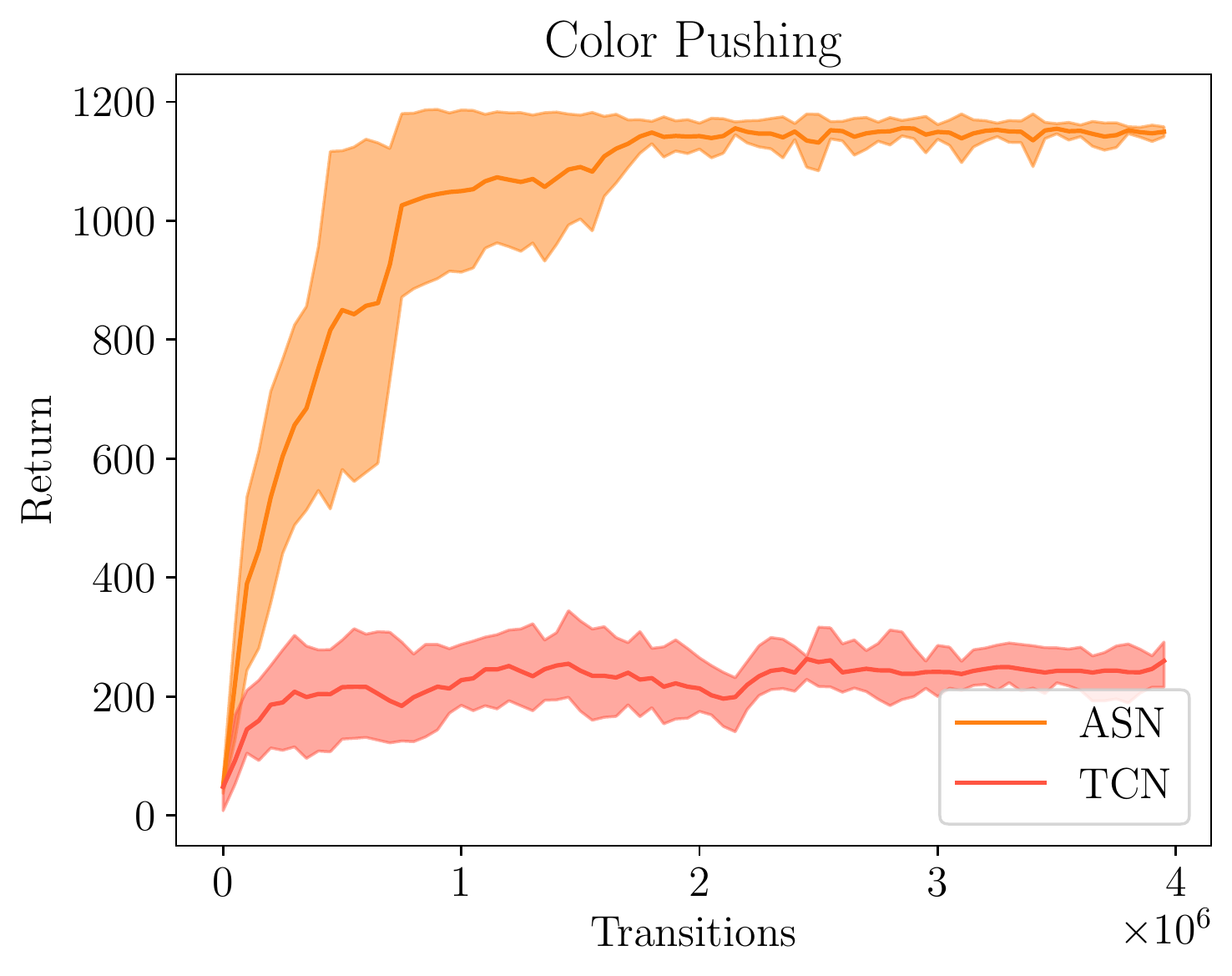}
    \includegraphics[width=0.49\linewidth]{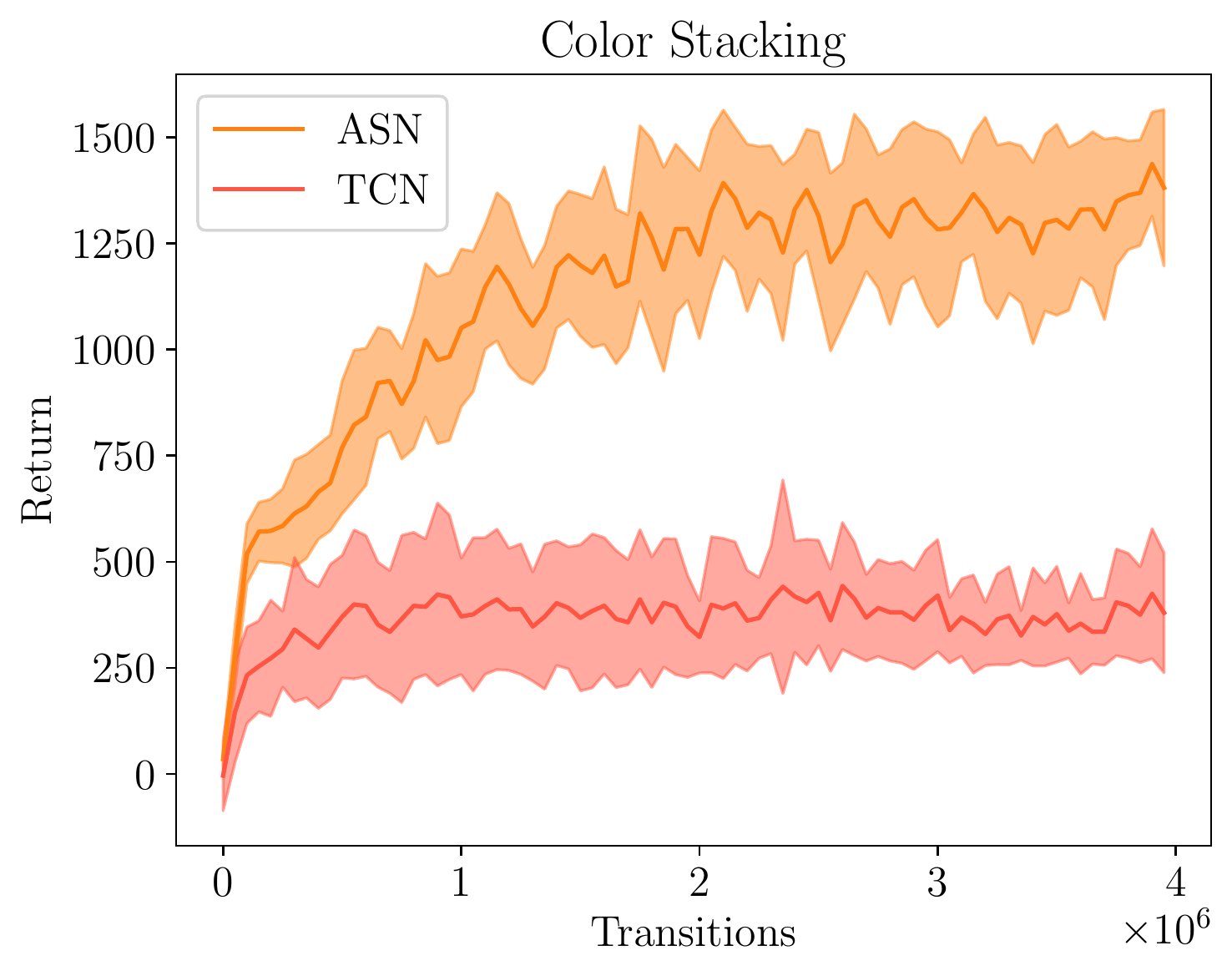}
    \caption{Results for training a continuous control policy with PPO on the unseen Color Pushing and Color Stacking tasks with the learned reward function. The plot shows mean and standard deviation over five training runs.}
    \label{fig:cp_learning_curve}
\end{figure}

Lastly, we integrate the learned metric within a RL-agent to imitate  an unseen task given a single video demonstration. Concretely, for learning  a continuous control policy on the color stacking (C) task we train the embedding on the tasks of two block stacking (A) and color pushing (B). Thus, successfully imitating the previously never seen color stacking task requires the interpolation of previously seen skills. Additionally, we also learn a continuous control policy for an unseen color pushing task, given an embedding trained on stacking and color stacking.
To train the agents, we use the distance measure in embedding space of the agent view $v^{t}_{a}$ and the demonstration frame  $v^{t}_{d}$ for timestep $t$ as the reward signal for the on-policy optimization algorithm PPO~\cite{ppo}: 
\begin{equation}
    r^{(t)} =
  \begin{cases}
    10-d\Big(E(v^{t}_{a}),E(v^{t}_{d})\Big) &  \text{if }~d\Big(E(v^{t}_{a}),E(v^{t}_{d})\Big)<\xi\\
    0  &  \text{otherwise,} 
  \end{cases}
\end{equation}
where $d$ is the euclidean distance and $\xi$ a constant threshold.
The agent state consists of the embedding $E(v^{t}_{a})$ and the joint angle of the robot. We train the policy with Adam and a learning rate of $10^{-5}$ and a batch size of 32.
To alleviate the problem of exploration and to focus on the quality of reward signal, we take random samples along the given demonstration as initial states
and reset the environment if the end effector vastly differs from the demonstration, following Peng \emph{et al.}~\cite{peng2018deepmimic}. 
The results are depicted in \Cref{fig:cp_learning_curve}.

Our method succeeds in solving the tasks, whereas the baseline TCN approach converges to a local minima. This demonstrates the effectiveness of our approach in reusing skills for a novel task given a single video demonstration.  Please note that training RL agents on tasks which have never been shown during the training of the embedding is very challenging, as it requires the discovery and reuse of task-independent skills.

Additionally, we evaluate the reward signal on an unseen color stacking task (C) for the real-world dataset. We plot a reward signal, which is based on the distance measurement of the task state for each timestep and a single goal frame from a different perspective, see Figure \ref{fig:rw}. The embedding of all models are trained on  tasks A, B and D.
To compare the different models we normalize the negative distance outputs for timestep between zero and one. The baseline model already give a similar reward for many initials states and goal states, despite the states being visually different. Our model shows a continuous and incremental reward as the task progresses and saturates as it is completed.

\section{Conclusion}
\label{sec:conclusion}

We proposed Adversarial Skill Networks, a model to leverage information from multiple label-free demonstrations in order to yield a meaningful embedding for unseen tasks. We showed that our approach is able to reuse learned skills for compositions of tasks and achieves state-of-the-art performance.  We demonstrate that the learned embedding enables training of continuous control policies to solve novel tasks that require the interpolation of previously seen skills.  Our results show that our model can find a good embedding for vastly different task domains. This is a first step towards discovery, representation and reuse of skills in the absence of a reward function.

Going forward, a natural extension of this work is the application of the learned distance metric in a real-world reinforcement learning setting and in environments that require a higher degree of interpolation for successful completion. Another promising direction for future work is the evaluation of the proposed approach in a sim-to-real setup~\cite{hermann2019adaptive}. 





\bibliographystyle{unsrt}

\bibliography{references}  

\end{document}